\documentclass[runningheads]{llncs}
\usepackage[utf8]{inputenc}
\DeclareUnicodeCharacter{00D7}{\(\times\)}
\usepackage[T1]{fontenc}
\usepackage{booktabs}
\usepackage{tabularx}
\usepackage{array}
\usepackage{enumitem} % nicer 
\usepackage{amsmath}
\usepackage{booktabs}
\usepackage{multirow}
\usepackage{graphicx}
\usepackage{xcolor}

\usepackage{hyperref}
\usepackage{xspace}
\usepackage{placeins}

\newcommand{\dataset}[0]{BioMedJImpact\xspace}

\begin{document}
\title{BioMedJImpact: A Comprehensive Dataset and LLM Pipeline for AI Engagement and Scientific Impact Analysis of Biomedical Journals}
\titlerunning{BioMedJImpact: Scientific Impact Analysis of Biomedical Journals}
\author{%
Anonymous Author(s)
% Ruiyu Wang\inst{1}\orcidID{0000-1111-2222-3333} \and
% Yuzhang Xie\inst{1}\orcidID{1111-2222-3333-4444} \and
% Xiao Hu\inst{1}\orcidID{2222-3333-4444-5555}\and
% Carl Yang\inst{1}\orcidID{2222-3333-4444-5555} \and
% Jiaying Lu\inst{1}\orcidID{0000-0001-9052-6951} 
}

%
% \authorrunning{Anonymous et al.}
\author{%
Ruiyu Wang \and
Yuzhang Xie \and
Xiao Hu \and
Carl Yang \and
Jiaying Lu
}

% 12 pages including references.
% Due: 11/15/2025
% \institute{%
% Anonymous Institution(s)
%Emory University, Atlanta, GA 30322, USA\\
%\email{\{jonathan.wang, yuzhang.xie, xiao.hu, j.carlyang, jiaying.lu\}@emory.edu}
% }
\institute{
Emory University, Atlanta, GA 30322, USA\\
\email{\{jonathan.wang, yuzhang.xie, xiao.hu, j.carlyang, jiaying.lu\}@emory.edu}
}

\maketitle
   % typeset the header of the contribution
%
\begin{abstract}
Assessing journal impact is central to scholarly communication, yet existing open resources rarely capture how collaboration structures and artificial intelligence (AI) research jointly shape venue prestige in biomedicine. We present \dataset, a large-scale, biomedical-oriented dataset designed to advance journal-level analysis of scientific impact and AI engagement. Built from 1.74 million PubMed Central articles across 2,744 journals, \dataset\ integrates bibliometric indicators, collaboration features, and LLM-derived semantic indicators for AI engagement. Specifically, the AI engagement feature is extracted through a reproducible three-stage LLM pipeline that we propose. Using this dataset, we analyze how collaboration intensity and AI engagement jointly influence scientific impact across pre- and post-pandemic periods (2016–2019, 2020–2023). Two consistent trends emerge: journals with higher collaboration intensity—particularly those with larger and more diverse author teams—tend to achieve greater citation impact, and AI engagement has become an increasingly strong correlate of journal prestige, especially in quartile rankings. To further validate the three-stage LLM pipeline we proposed or deriving the AI engagement feature, we conduct human evaluation, confirming substantial agreement in AI relevance detection and consistent subfield classification. Together, these contributions demonstrate that BioMedJImpact serves as both a comprehensive dataset capturing the intersection of biomedicine and AI, and a validated methodological framework enabling scalable, content-aware scientometric analysis of scientific impact and innovation dynamics. Code is available at \url{https://github.com/JonathanWry/BioMedJImpact}.
%The abstract should briefly summarize the contents of the paper in 150--250 words.

\end{abstract}

\section{Introduction}

The scientific impact of journals plays a central role in academic communication, influencing research visibility, funding allocation, and institutional evaluation~\cite{garfield1955citation}. In biomedicine, metrics such as the impact factor (IF), citation counts, and journal ranking (e.g., JCR quartiles, SCImago Journal Rank) serve as key decision-making tools for authors, institutions, and funders~\cite{dong2005_if_revisited}. These indicators are sensitive to changes in the research landscape. For example, many general medical journals such as \textit{The New England Journal of Medicine} and \textit{The Lancet} saw sharp spikes in IF in 2021 due to the COVID-19 publication surge, followed by a return toward pre-pandemic levels in 2022~\cite{kim2024_covid_if}. Meanwhile, artificial intelligence (AI) has increasingly transformed biomedical research, including genomic prediction~\cite{alharbi2022_deeplearning_genomics} and clinical imaging~\cite{zhou2020_medimg}, reshaping how research is conducted and evaluated, and potentially altering traditional scientific impact indicators.

Existing open datasets on scientific impact (e.g., AMiner~\cite{tang2008_aminer}, DBLP~\cite{ley2002_dblp}, and Microsoft Academic Graph~\cite{wang2020_mag}) have greatly advanced large-scale analyses of scholarly networks and citation behavior; DBLP focuses on computer science, while AMiner and MAG provide broad, cross-disciplinary coverage. However, these existing resources are not designed specifically for the biomedical domain, and they lack the granularity needed to capture AI’s influence within biomedical research. 
To tackle these problems, we leverage multi-source data to build a new dataset on biomedical journals' scientific impact, named as \dataset. \dataset\ integrates three major categories of features:
(a) Bibliometric indicators (e.g., impact metrics, citation counts),
(b) Collaboration indicators (e.g., Author diversity, institutional diversity), and
(c) AI-related indicators(e.g., AI engagement rate, AI subfield distributino) that quantify the presence and distribution of AI-related research across journals. Specifically, bibliometric indicators are sourced from the Journal Citation Reports (JCR) and CiteFactor; collaboration indicators are derived from PubMed Central (PMC) metadata, capturing author and institutional structures; and the AI-related indicators are derived from article abstracts using a large language model (LLM)-based pipeline, enabling us to systematically identify AI-related publications and their associated subfields. 

In total, we build \dataset, a comprehensive dataset for analyzing biomedical journals' scientific impact, which consists of 2,744 journals. We derive 55 comprehensive features, covering Bibliometric indicators, collaboration indicators, and AI-related indicators. Based on PMC data, after matching all journals by source and publication year, we identified 1,740,112 papers, which were analyzed using our LLM-based pipeline, ending with an overall AI engagement rate of 3.77\%. We further conduct correlation analysis and identify 26 significant factors. Collaboration intensity—particularly larger and more diverse author teams—shows a consistent positive association with citation impact, while AI engagement rate was shown as an indicator of journal prestige. Although its influence was less stable in 2019, by 2023 higher AI engagement rates were strongly aligned with higher quartile rankings. Together, these findings highlight how \textbf{\dataset} combines LLM-derived semantic indicators with traditional bibliometric and collaboration indicators, offering a unified and scalable framework for understanding the evolving relationship between content, collaboration, and scientific impact in biomedical publishing.
\section{Related Work}
\subsection{Scientific Impact Modeling}
% please refer to https://dl-acm-org.proxy.library.emory.edu/doi/pdf/10.1145/2684822.2685314

% we can summarize them as "descriptive metrics" derived from citations?
Scientific impact has long been a core concern in scientometrics and information science, offering insight into how scholarly influence accumulates and providing practical tools for evaluating research quality, allocating funding, and guiding publication strategies~\cite{garfield1955citation}. Among various indicators, citations remain the primary quantitative signal, forming the basis of a family of \emph{citation-based indicators} used to assess journals, authors, and individual papers~\cite{bornmann2008_citationreview}. At the journal level, Garfield’s journal impact factor (IF) formalized citation aggregation as a venue-level indicator~\cite{garfield2006_jifmeaning}, while the Journal Citation Reports (JCR) quartile scheme (Q1--Q4) situates journals within disciplinary hierarchies, providing a coarse yet actionable measure of prestige for authors, editors, and institutions.~\cite{krampl2019jcr}. Despite well-documented limitations including field-normalization challenges and citation skewness, these citation-based indicators remain interpretable and comparatively stable benchmarks that correlate with long-term scientific attention~\cite{wang2013_science}.
% focus on other factors (other than citation): collaboration. content/topic -> AI engagemtnt 
Beyond citation-based indicators, extensive research has explored how different factors contribute to scientific impact. Collaborative indicators such as team size and co-authorship networks have been shown to correlate with citation influence across disciplines \cite{lariviere2015_teamsize,wuchty2007_science}. Studies also indicate that thematic and linguistic content learned from titles and abstracts have been found to encode meaningful cues of scholarly influence \cite{cohan2020_specter}. Motivated by these findings, we construct a biomedical-focused, open dataset that integrates journal-level bibliometric indicators and collaboration indicators derived from author and affiliation metadata. 

\subsection{Large Language Model based Feature Extraction}
% this serve as strong connection to PAKDD LLM track

% para1: advances on LLM for feature extraction; find citation on how works have been utilizing llm for feature extraction
Alongside traditional feature extraction methods, recent advances in large language models (LLMs) have fundamentally changed how features can be mined from scientific papers. Conventional feature extraction pipelines typically rely on handcrafted rules or supervised NLP models that are expensive to develop and maintain due to annotation costs, domain drift, and ongoing schema adaptation~\cite{wadden2020_entity_event_review}. By contrast, LLMs enable prompt-based extraction that can screen documents for topical relevance, identify domain-specific terms, and map those terms to controlled taxonomies. Recent surveys document strong zero-shot and few-shot performance of LLMs for generative information extraction in broad, domain-general setting such as named-entity, relation, and event extraction~\cite{xu2024_generativeIE_springer}.
% para2: connect to LLM based feature extraction for scientific impact factor mining. Especially focusing on biomedical journals.
In biomedical corpora, LLMs have been applied to instruction–following information extraction across core tasks including named–entity recognition, relation extraction, and procedure extraction~\cite{tran2024_bioinstruct}. In concrete biomedical applications, LLMs have likewise demonstrated practical utility. In radiology, the RadEx benchmark uses prompted LLMs to convert free-text reports into structured tuples, extracting findings, anatomical sites, and modifiers~\cite{reich2024_scoping_radiologyIE}. Similarly, LLMs have demonstrated their feasibility, accuracy, and efficiency for large-scale study design elements (PICO) extraction from clinical abstracts in PubMed~\cite{reason2024_pico_genai}. In this study, we build an LLM pipeline that (i) screens abstracts for AI relevance, (ii) extracts and validates keyword mentions, and (iii) maps them to a controlled AI-subfield taxonomy. 

\section{Dataset Construction}

\subsection{Multisource Integration for Initial Dataset Construction}

In this study, we curate \dataset, a comprehensive journal-level dataset for large-scale analysis of biomedical journal impact and AI engagement, by integrating data from:
(i) the \textit{PubMed Central Open Access subset} (PMC-OA)~\cite{PMC-OA} for full-text and metadata of biomedical articles,
(ii) \textit{Journal Citation Reports} (JCR)~\cite{krampl2019jcr} for journal bibliometric records including historical impact metrics and citation information, and 
(iii) the \textit{Directory of Open Access Journals} (DOAJ)~\cite{morrison2017doaj} for journal-level open-access policies and publication practices.
Based on these resources, we assemble 17 per–journal, per–year indicators (see Table~\ref{tab:feature-summary} for details). Among all data sources, PMC-OA serves as the core foundation of \dataset. It provides full-text and metadata for \textbf{4,298} biomedical journals. After matching these journals with available bibliometric records from JCR, we retain \textbf{2,744} journals for downstream analysis of content, citation patterns, and collaboration indicators. Of these, \textbf{1,694} journals are indexed in the DOAJ, enabling the integration of open-access policies and publication practices into the dataset.
The finalized version of \dataset\ will be released to the research community upon acceptance to promote transparency, reproducibility, and further investigation into biomedical journal impact.

\subsubsection{Dataset/Year Split.} 
To facilitate downstream modeling and isolate temporal effects, we partition the unified dataset into two temporal subsets: \textbf{BioMedJImpact~2019} (2016–2019) and \textbf{BioMedJImpact~2023} (2020–2023). Within each subset, we focus on three commonly used journal-level targets: \emph{Impact Factor}, \emph{Quartile}, and \emph{Total Cites (3Y)}. These targets are widely used in academic assessment systems: Impact Factor reflects short-term citation influence, Quartile indicates a journal’s relative standing within its subject category, and Total Cites (3Y) captures sustained citation accumulation.
The temporal split is motivated by structural shifts in publishing behavior and citation dynamics during the COVID-19 period. Separating pre- and post-pandemic data helps ensure that observed relationships are not confounded by pandemic-related disruptions.
For each subset, we retain only journals with a valid Impact Factor in the subset’s target year. After filtering, \textbf{BioMedJImpact~2019} contains 1{,}367 journals and \textbf{BioMedJImpact~2023} contains 2{,}685 journals. In terms of coverage, bibliometric completeness remains high: over 90\% of journals include quartile rankings and citation-based metrics (Total Cites (3Y)). Table~\ref{tab:dataset-summary} summarizes the retained sets.

% learn from this table: https://www.nature.com/articles/s41597-022-01899-x/tables/2 

\begin{table}[t]
\vspace{-0.2cm} 
\centering
\small
\caption{Summary of statistics for the \dataset-\textbf{2019} and -\textbf{2023} subsets.} 
%The “Target Year” refers to the year in which Impact Factor (IF), Quartile, and three-year citation counts (\emph{Total Cites\_3Y}) are used as outcome variables.
%\emph{Std} denotes standard deviation.}
\label{tab:dataset-summary}
\begin{tabular}{lcc}
\toprule
Statistic \textbackslash{} Sub Dataset & -2019 & -2023 \\
\midrule
\# Journals & 1367 & 2685 \\
\hline
\# Journals with Quartile & 1243 & 2321 \\
\quad Percentage of Q1 journals & 57.2\% & 46.92\% \\
\hline
\# Journals with IF & 1367 & 2685 \\
\quad Avg IF & 3.43 & 3.35\\
\quad Std IF & 3.12 & 4.01 \\
\hline
\# Journals with Total Cites\_3Y & 1247 & 2314 \\
\quad Avg Total Cites (3Y) & 27263 & 33018\\
\quad Std Total Cites (3Y) & 116497 & 132542 \\
\bottomrule
\end{tabular}
\vspace{-0.3cm} 
\end{table}

\begin{figure*}[htbp!]
\vspace{-0.3cm} 
\centering
\begin{tabular}{cc}
    % -------- Row 1 --------
    \includegraphics[width=0.35\textwidth]{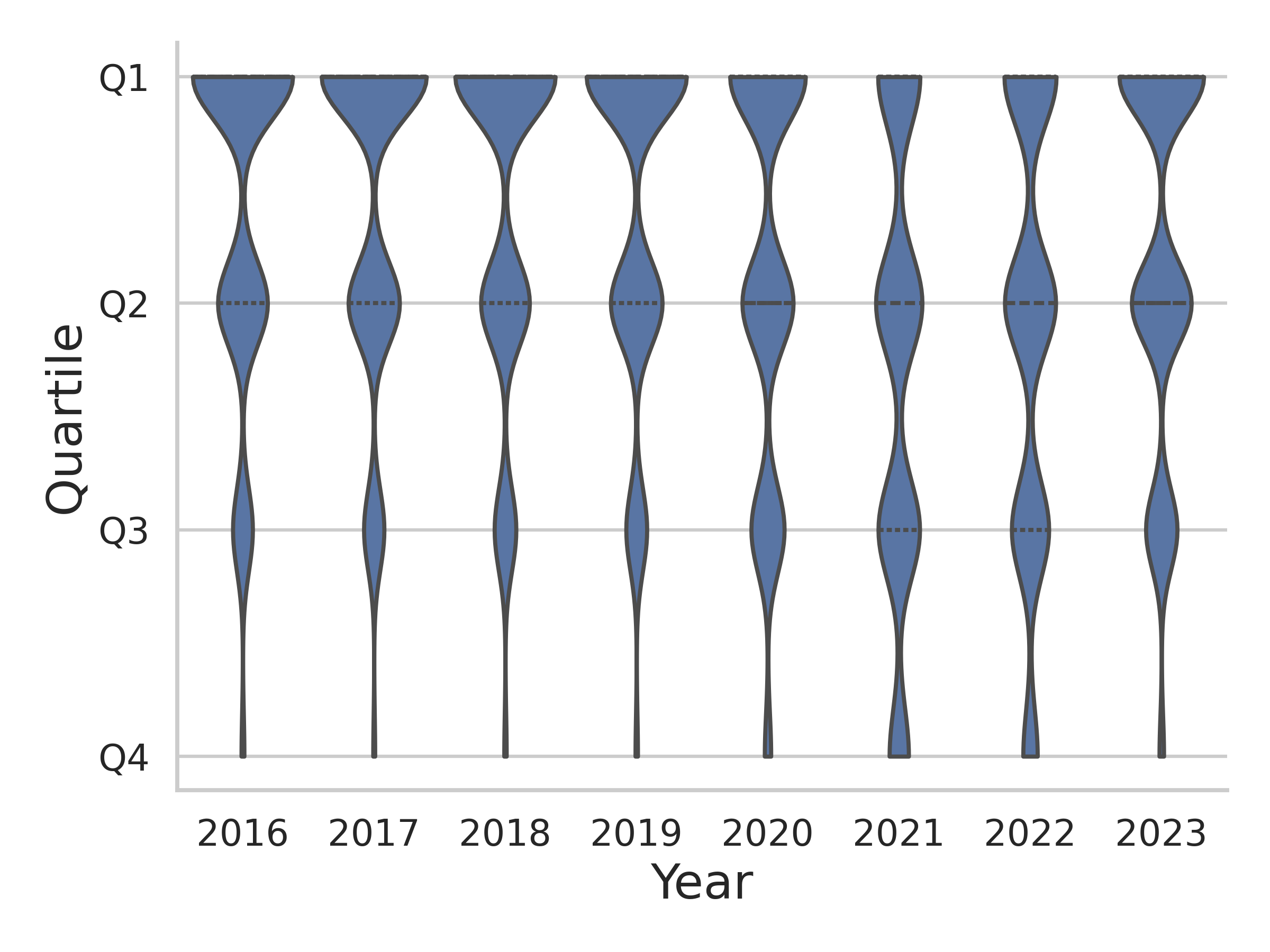} &
    \includegraphics[width=0.35\textwidth]{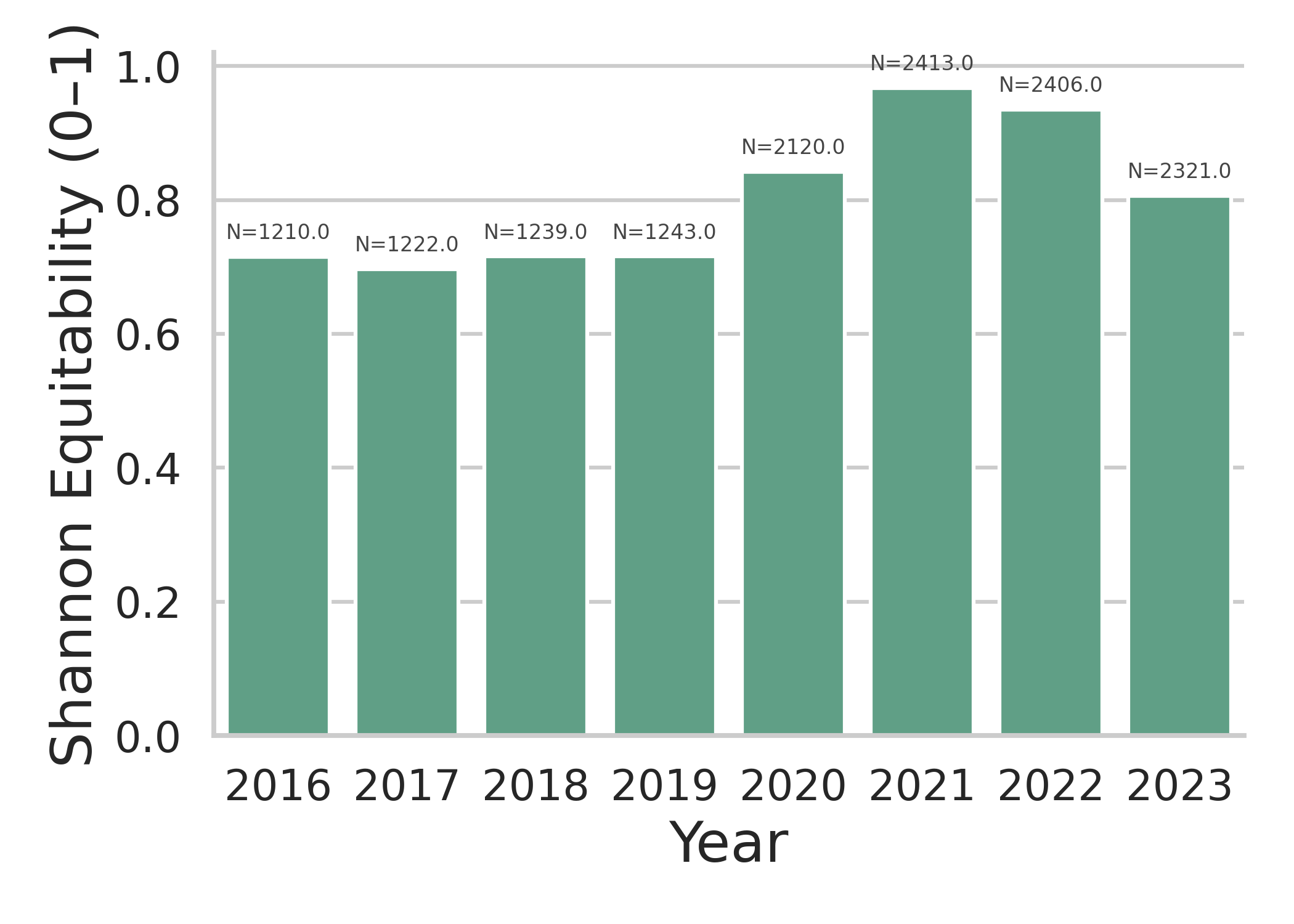} \\[-2pt]
    (a) Quartile distribution & (b) Quartile balance  \\[6pt]

    % -------- Row 2 --------
    \includegraphics[width=0.35\textwidth]{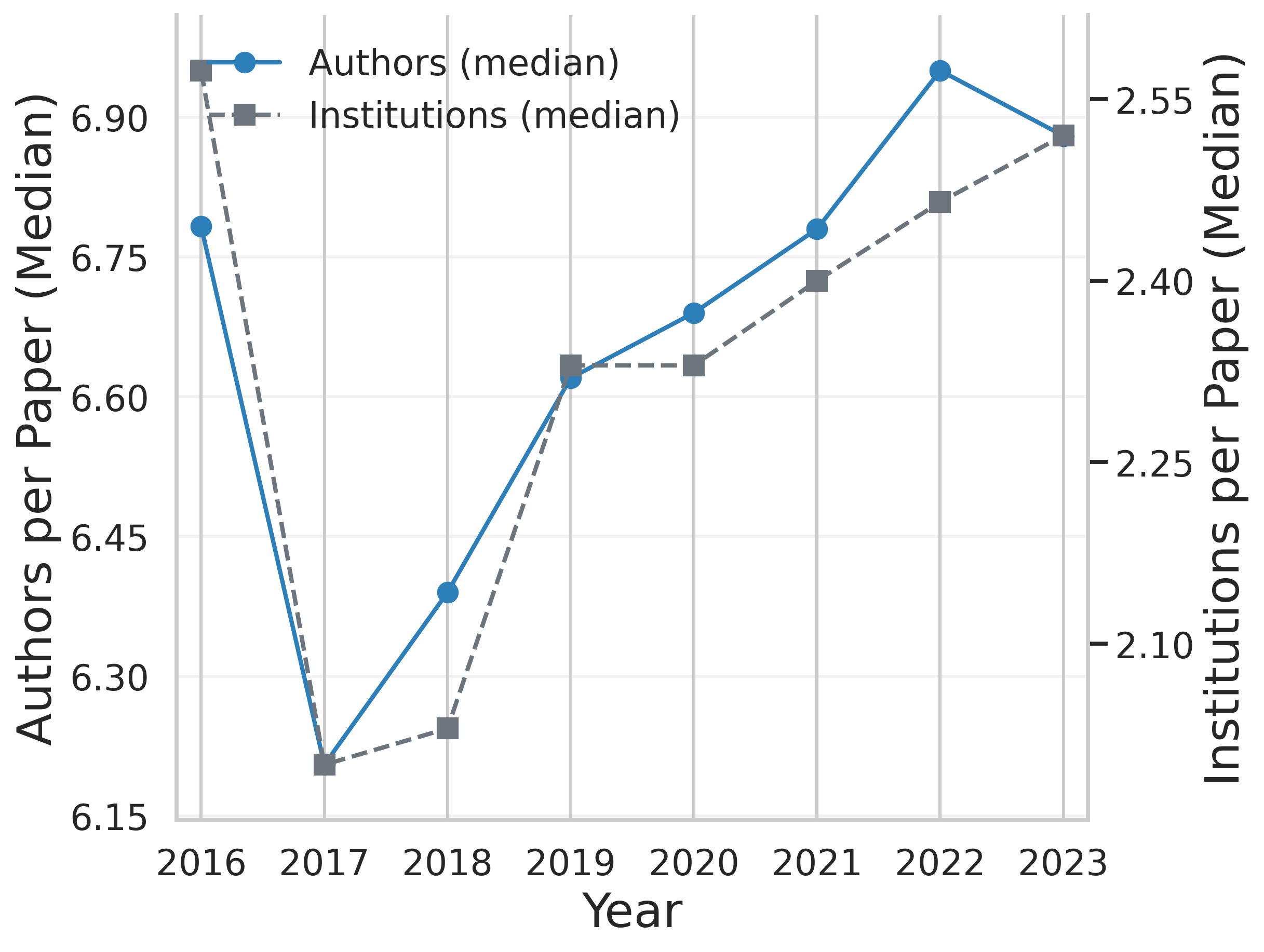} &
    \includegraphics[width=0.35\textwidth]{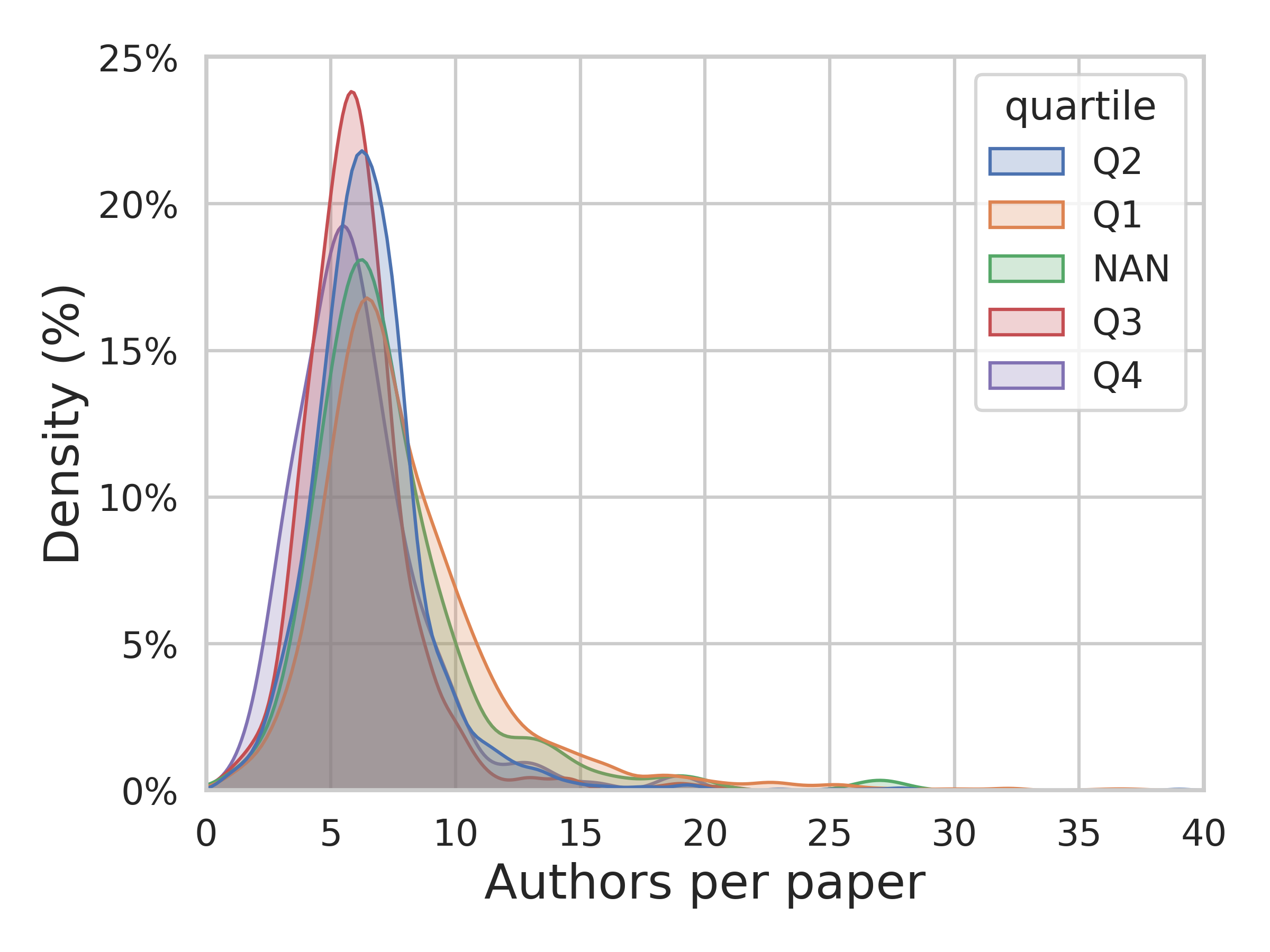} \\[-2pt]
    (c) Collaboration trends & (d) Authors per paper by quartile
\end{tabular}
\caption{Exploratory visualizations of quartile and collaboration indicators in the integrated dataset. Panels (a)–(b) summarize quartile dynamics and stability; panels (c)–(d) depict collaboration trends and collaboration intensity by quartile.}
\label{fig:quartile-collaboration-grid}
\vspace{-0.5cm} 
\end{figure*}

\subsubsection{Bibliometric Indicators.}
\dataset integrated bibliometric indicators from multiple publicly available sources. Historical journal indicators are collected from JCR hosted on ResearchGate\footnote{\href{https://www.researchgate.net/}{https://www.researchgate.net/}} (2016–2024), and missing values were supplemented using CiteFactor\footnote{\href{https://www.citefactor.org}{https://www.citefactor.org}}
. Extracted fields include journal title, ISSN/EISSN, subject category, quartile ranking (Q1--Q4), impact factor, and total citations. Policy attributes are integrated via cross-referencing with the DOAJ, which provides publication delay (in weeks), author copyright-retention status, and article processing charges. All sources are harmonized by ISSN/EISSN as unique identifiers, with fuzzy title matching applied for unresolved cases.

\subsubsection{Collaboration Indicators.}
To characterize author collaboration patterns, we process full-text XML archives from PMC, a free full-text repository maintained by the U.S.\ National Institutes of Health’s National Library of Medicine. PMC provides rich article-level metadata, including author affiliations, article types, and author-supplied keywords. From each article, we extract the number of distinct institutions and participating countries using both structured tags (\texttt{<institution>}, \texttt{<country>}) and fallback string-pattern matching when such tags are absent. 
For each journal--year pair, we compute summary statistics that capture the structure of author collaboration indicators, including the mean, standard deviation, and interquartile range of authors and institutions per article. We additionally define a \emph{cross-country collaboration rate} as the proportion of articles with author affiliations spanning multiple countries. These metrics enable standardized comparisons of institutional and international collaboration intensity across disciplines and temporal spans.

\subsubsection{Descriptive Insights.}
We further conduct an exploratory analysis to examine structural and temporal variation in bibliometric indicators and author collaboration indicators. Specifically, we assess (1) the longitudinal stability of journal impact distributions and (2) the evolution of collaborative practices. Figure~\ref{fig:quartile-collaboration-grid} presents a four-panel overview summarizing these patterns across three analytical dimensions. 
Figure~\ref{fig:quartile-collaboration-grid}(a,b) depict the temporal distribution of journal quartiles from 2016 to 2023. As shown in figure~(a) The overall composition remains relatively stable, with most journals occupying the mid-tiers (Q2--Q3) and only minor inter-annual variation. The \emph{Shannon equitability index} in figure~(b) increases modestly during 2021--2022, indicating a temporary phase of greater balance in quartile representation before returning to prior levels by 2023.  
Figures~\ref{fig:quartile-collaboration-grid}(c,d) show longitudinal trends in author and institutional academic collaboration. As shown in figure~(c), the median number of authors per paper decreases sharply in 2017 before increasing steadily through 2022, accompanied by a parallel rise in the number of contributing institutions. Figure~(d) compares the distribution of authors per article across quartiles, revealing substantial overlap among tiers; this suggests that collaboration intensity, while increasing over time, is not itself a strong determinant of journal ranking.

% Sec 3.2
% merged into data construction
\subsection{LLM-Based Feature Enrichment for Journal AI Engagement}

%\jiaying{we need to add one paragraph stating this LLM pipeline leads to a more comprehensive dataset.}

Besides traditional bibliometric and collaboration indicators, we further enrich \dataset\ with features derived from a LLM–based extraction pipeline. This enrichment extends the dataset beyond structural metadata to include content-level indicators that quantify each journal’s engagement with artificial intelligence (AI)–related research. By integrating these semantic features with previously extracted collaboration indicators, the resulting dataset supports a more comprehensive set of predictors spanning structural, behavioral, and topical dimensions.
Table~\ref{tab:ai_completeness} summarizes the coverage of these indicators. “ANY AI-eng.” and “ANY collaboration” indicate journals with at least one corresponding indicator in any of the three preceding years, whereas “FULL AI-eng. (3Y)” requires that all AI engagement indicators are present in each of the three years. Consistent with the definitions in the caption, around 900–1{,}000 journals in each period contain at least one valid AI or collaboration feature, reflecting both broad coverage and the increasing availability of LLM-derived AI engagement rate over time.
The full code for the LLM-based AI extraction pipeline is publicly available at \url{https://github.com/JonathanWry/BioMedJImpact}
.

\begin{table}[t]
\vspace{-0.2cm} 
\caption{Dataset completeness summary. 
“ANY AI-eng.” counts journals with at least one AI engagement indicator in any of the three years. 
“FULL AI-eng. (3Y)” requires all AI engagement indicators to be present for all three years. 
“ANY collaboration indi.” counts journals with at least one collaboration indicator.}
\centering
\small
\begin{tabular}{lcc}
\toprule
Statistic & 2019 & 2023 \\
\midrule
\# Journals w/ ANY AI-eng & 902 & 1010 \\
\# Journals w/ FULL AI-eng. (3Y) & 631 & 803 \\
\# Journals w/ ANY collaboration indi. & 1095 & 979 \\
\bottomrule
\end{tabular}
\label{tab:ai_completeness}
\vspace{-0.3cm} 
\end{table}

\begin{table}[ht!]
\vspace{-0.4cm} 
\caption{Summary of features. Feature\textsubscript{Y-1}, \textsubscript{Y-2}, and \textsubscript{Y-3} denote covariates from one to three years prior to the prediction year.}
\label{tab:feature-summary}
\centering
\small
\resizebox{\columnwidth}{!}{%
\begin{tabular}{p{0.4\columnwidth} p{0.6\columnwidth}}
\toprule
\textbf{Feature Group} & \textbf{Features Included} \\
\midrule
\textbf{Bibliometric Indicators} 
& \textbullet\ Impact Factor \textsubscript{Y-1}, \textsubscript{Y-2}, \textsubscript{Y-3} \newline
% \textbullet\ Category \newline
\textbullet\ Quartile \textsubscript{Y-1}, \textsubscript{Y-2}, \textsubscript{Y-3} \newline
\textbullet\ Total Cites (3Y) \textsubscript{Y-1}, \textsubscript{Y-2}, \textsubscript{Y-3} \newline
\textbullet\ Total References \textsubscript{Y-1}, \textsubscript{Y-2}, \textsubscript{Y-3} \newline
\textbullet\ Publication Count \textsubscript{Y-1}, \textsubscript{Y-2}, \textsubscript{Y-3}  \newline
\textbullet\ Publication Delay (in weeks) \newline
\textbullet\ Author Copyright Retention \newline
\textbullet\ Article Processing Charges \newline
\textbullet\ Subject Category \\ 
\midrule
\textbf{Collaboration Indicators} 
& \textbullet\ Avg. Authors \textsubscript{Y-1}, \textsubscript{Y-2}, \textsubscript{Y-3} \newline
\textbullet\ Std. Authors \textsubscript{Y-1}, \textsubscript{Y-2}, \textsubscript{Y-3} \newline
\textbullet\ Author Quartiles \textsubscript{Q25, Q50, Q75 × Y-1, Y-2, Y-3} \newline
\textbullet\ Avg. Institutions \textsubscript{Y-1}, \textsubscript{Y-2}, \textsubscript{Y-3} \newline
\textbullet\ Std. Institutions \textsubscript{Y-1}, \textsubscript{Y-2}, \textsubscript{Y-3} \newline
\textbullet\ Institution Quartiles \textsubscript{Q25, Q50, Q75 × Y-1, Y-2, Y-3} \newline
\textbullet\ Cross-country collaboration rate \textsubscript{Y-1}, \textsubscript{Y-2}, \textsubscript{Y-3} \\
\midrule
\textbf{AI-Related Indicators} 
& \textbullet\ AI Engagement percentage \textsubscript{Y-1}, \textsubscript{Y-2}, \textsubscript{Y-3} \\
% \textbullet\ Level 1: Subdomain distribution \textsubscript{Y-1}, \textsubscript{Y-2}, \textsubscript{Y-3} \newline
% \textbullet\ Level 2: Fine-grained subfield distribution \textsubscript{Y-1}, \textsubscript{Y-2}, \textsubscript{Y-3} \\
\bottomrule
\end{tabular}
}
\vspace{-0.6em}
\end{table}

\subsubsection{LLM-Based Content Analysis on AI from PMC.}
\begin{figure*}[t]
  \centering
  \vspace{-0.3cm} \includegraphics[width=0.90\textwidth]{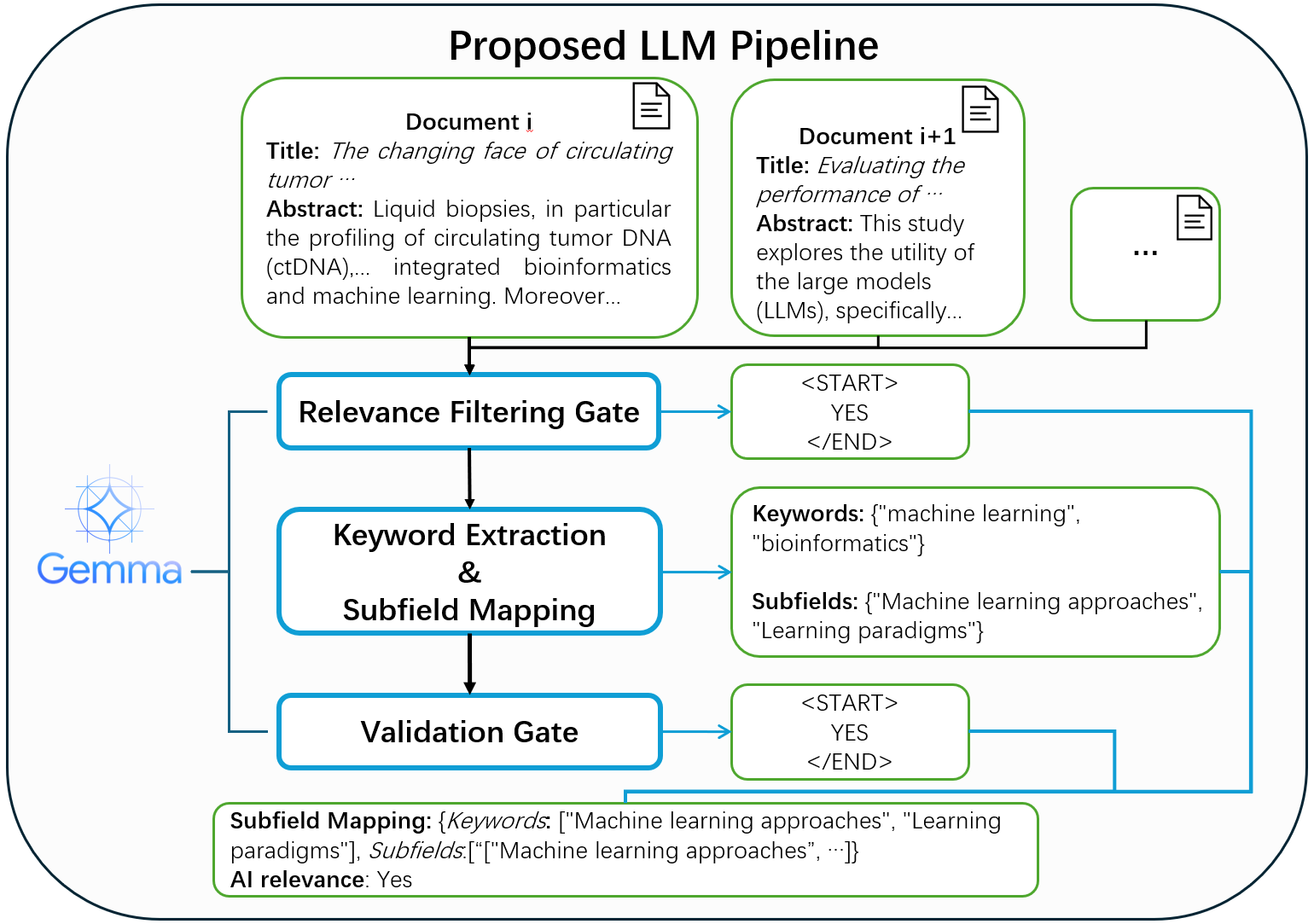}
  \caption{Overview of our LLM pipeline for AI engagement analysis from PMC abstracts. Step~1 filters AI-relevant abstracts. Step~2 extracts AI terms and maps them to a controlled taxonomy of AI subfield. Step~3 validates extracted terms to reduce ambiguity and false positives.}
  \label{fig:llm-pipeline}
  \vspace{-0.5cm} 
\end{figure*}

To analyze AI involvement and related AI thematic content information, we perform large-language-model–based annotation of PMC article abstracts to estimate journal-level engagement with AI research, which is described in Figure \ref{fig:llm-pipeline}. Using vLLM with the Gemma-3-12B-IT model, we implement a three-step pipeline: 
\vspace{-0.2cm}
\begin{enumerate}
  \item \textbf{Relevance Filtering Gate:} Each abstract is first screened by a LLM classification prompt that determines whether it is explicitly relevant to artificial intelligence or machine learning.  For instance, abstracts containing phrases such as “deep learning–based model,” “AI-assisted diagnosis,” or “neural network training” are labeled as AI-relevant. Non-technical mentions (e.g., “intelligent design”) are filtered out.
  
  \item \textbf{Keyword Extraction and Subfield Mapping:} Abstracts identified as AI-relevant are then processed by a second LLM prompt that simultaneously (1) extracts AI-related keywords (e.g., CNN, transformer, reinforcement learning, image segmentation) and (2) maps each abstract to one or more predefined AI subfield, including \emph{Natural Language Processing}, \emph{Computer Vision}, \emph{Learning Algorithms}, \emph{Knowledge Representation}, \emph{Search}, and \emph{Distributed AI}. This integrated keyword–subfield reasoning step enables consistent subfield assignment and supports downstream analysis of AI research themes.
  
  \item \textbf{Validation Gate:} A secondary verification prompt re-evaluates all extracted keywords to confirm their alignment with AI subfields and removes ambiguous or noisy terms (e.g., “training session” or “learning curve”). This ensures semantic precision and minimizes false positives in downstream statistical analyses.
\end{enumerate}
\vspace{-0.2cm}

\noindent From this pipeline, we derive two principal features:  
(1) the \emph{AI engagement rate} that states proportion of articles within each journal–year flagged as AI-related:
\begin{equation}
E_{j,t} = \frac{N^{\text{AI}}_{j,t}}{N^{\text{total}}_{j,t}},
\end{equation}
where \(N^{\text{AI}}_{j,t}\) is the number of AI-related abstracts in journal \(j\) during year \(t\), and \(N^{\text{total}}_{j,t}\) is the total number of abstracts published by that journal in the same year.
(2) the \emph{AI subfield distribution}, summarizing the relative composition of AI subfields for each journal–year:
\begin{equation}
C_{j,t,k} = \frac{N^{\text{AI}}_{j,t,k}}{\sum_{k} N^{\text{AI}}_{j,t,k}},
\end{equation}
\noindent where \(N^{\text{AI}}_{j,t,k}\) denotes the number of AI-related abstracts in subject category \(k\). This distribution quantifies the proportion of engagement across AI subfields (e.g., NLP, Computer Vision, etc.) within each journal–year.

\subsubsection{AI Engagement Patterns}

\begin{figure*}[htbp!]
\vspace{-0.5cm} 
\centering
\begin{tabular}{cc}    \includegraphics[width=0.49\textwidth]{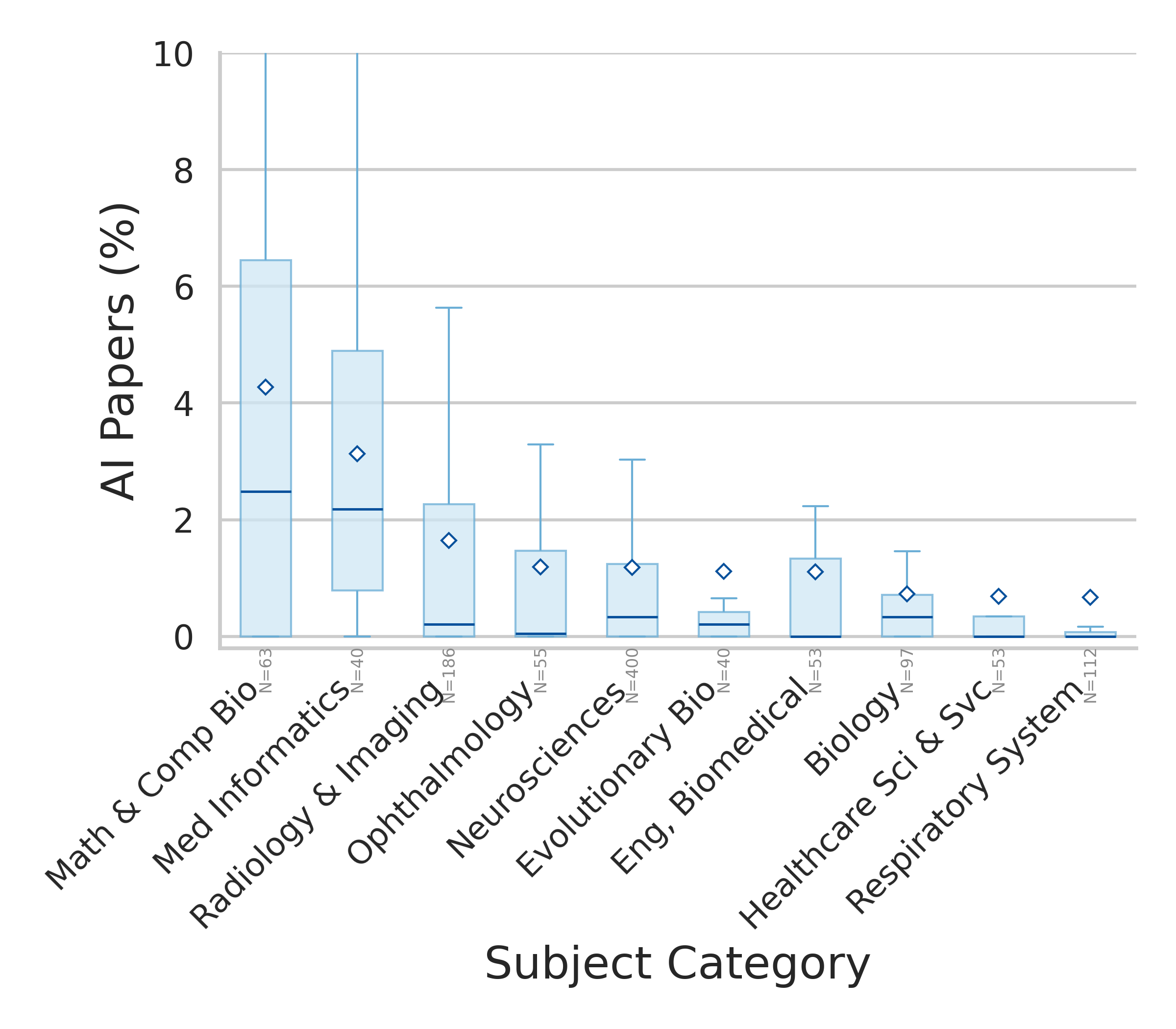} &
    \includegraphics[width=0.49\textwidth]{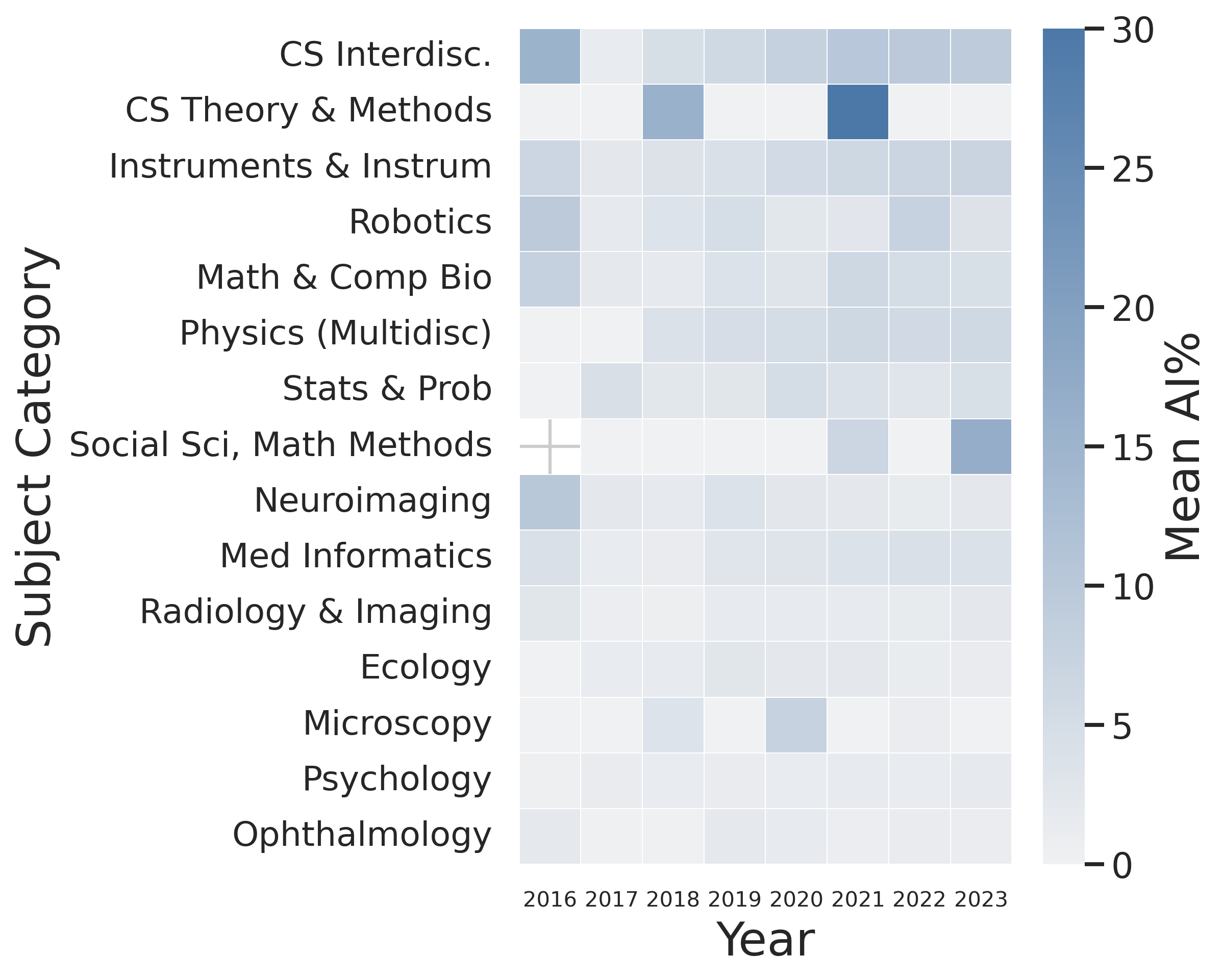} \\[-2pt]
    (a) Pooled AI\% by subject category & (b) AI\% by year and subject category
\end{tabular}
\caption{AI engagement patterns derived from LLM-based content annotation. Panel (a): Top‑10 by pooled mean AI\% over all journal–year rows within each subject category; boxes show distributions, diamonds show means. Panel (b): Top‑15 by year‑normalized mean AI\%—mean across journals within each category–year, then mean across years.}
\label{fig:ai-engagement-grid}
\vspace{-0.5cm}
\end{figure*}

Figures~\ref{fig:ai-engagement-grid}(a,b) summarize AI-related publication trends based on LLM-derived annotations. Figure~(a) shows that AI mentions are most frequent in multidisciplinary fields, especially those linked to computer science and mathematics. High engagement levels also appear in imaging-focused domains such as \emph{Radiology and Medical Imaging}, as well as in \emph{Neuroscience}, where machine learning is commonly applied to structured, high-dimensional data. Figure~(b) displays a temporal heatmap of AI activity across the top 30 subject categories (ranked by mean AI proportion). Most fields exhibit a steady year-over-year increase in AI-related content. Notably, \emph{Robotics} shows a distinct upward trajectory beginning in the mid-2010s, reflecting expanded use of AI methods in that discipline.

\section{Correlation Analysis}
\label{sec:dataset-correlation}
\FloatBarrier

Building on the descriptive summaries above, we next assess feature interdependencies and their predictive relevance. We study associations between lagged journal features and three outcomes--\emph{Impact Factor (IF)}, \emph{Total Cites (3Y)}, and \emph{Quartile}--using linear mixed--effects models with random intercepts by subject category~\cite{pinheiro2000mixed}. Let \(j\) index journals, \(c(j)\) denote the subject category of journal \(j\), and \(t\) denote the prediction year (2019 or 2023). For each outcome \(y_{j,t}\), we fit
\begin{equation}
\label{eq:mixedlm}
y_{j,t}
= \alpha
+ \mathbf{x}^{\top}_{j,t-1:t-3}\boldsymbol{\beta}
+ u_{c(j)} + \varepsilon_{j,t},\quad
u_{c} \sim \mathcal{N}(0,\tau^{2}),\;\;
\varepsilon_{j,t} \sim \mathcal{N}(0,\sigma^{2}),
\end{equation}
where \(y_{j,t}\) denotes the target outcome for journal \(j\) in year \(t\), \(\alpha\) being the global intercept. The vector \(\mathbf{x}_{j,t-1:t-3}\) contains the lagged covariates from the three years preceding citation-based indicators\(t\), including publication counts, reference counts, open-access status, collaboration indicators, and AI engagement rates. We deliberately exclude contemporaneous features at year \(t\) to prevent target leakage. The fixed-effect vector \(\boldsymbol{\beta}\) captures within-subject category partial associations. The term \(u_{c(j)}\) is a random intercept associated with the subject category \(c(j)\), assumed to follow \(u_{c} \sim \mathcal{N}(0, \tau^2)\), which captures persistent field-level differences not explained by covariates (e.g., radiology vs.\ oncology). The idiosyncratic error term \(\varepsilon_{j,t}\) is assumed to follow \(\mathcal{N}(0, \sigma^2)\), independently across journals and years.

\begin{table}[!htbp]
 \caption{Linear mixed–effects summary for \textbf{\dataset~2019}. Significant predictors only; robust Std. Errors in parentheses.}
\centering
\setlength{\tabcolsep}{6pt}
\resizebox{\columnwidth}{!}{%
\begin{tabular}{l l r r c c}
\toprule
\textbf{Target} & \textbf{Variable} & \textbf{Coef.} & \textbf{Std. Err.} & \textbf{95\% CI (L--H)} & \textbf{Signif.} \\
\midrule
\multirow{6}{*}{Impact Factor}
 & Avg\_Authors\_2016      & 0.808   & 0.085   & [0.641, 0.975]           & *** \\
 & Total\_Refs\_2016       & -1.00e-5 & 3.00e-6 & [-1.90e-5, -5.00e-6]     & *** \\
 & AI\_Perc\_By\_LLM\_2018 & 14.0    & 4.44    & [5.33, 22.7]             & **  \\
 & Std\_Institutions\_2018 & -0.408  & 0.129   & [-0.661, -0.154]         & **  \\
 & Std\_Authors\_2016      & -0.056  & 0.018   & [-0.091, -0.021]         & **  \\
 & Total\_Refs\_2017       & 1.80e-5 & 7.00e-6 & [3.00e-6, 3.20e-5]       & *   \\
\midrule
\multirow{10}{*}{Total Cites}
 & publication\_count\_2016 & -159    & 14.7    & [-187, -130]             & *** \\
 & publication\_count\_2018 &  158    & 15.3    & [128, 188]               & *** \\
 & publication\_count\_2017 &   96.1  & 18.9    & [59.1, 133]              & *** \\
 & Total\_Refs\_2018        &  -0.878 & 0.183   & [-1.24, -0.520]          & *** \\
 & Total\_Refs\_2017        &   0.931 & 0.278   & [0.390, 1.48]            & *** \\
 & Std\_Institutions\_2017  &  1.03e4 & 3.48e3  & [3.51e3, 1.72e4]         & **  \\
 & Author\_Copyright\_Retention & -2.28e4 & 7.73e3 & [-3.80e4, -7.70e3]   & **  \\
 & Std\_Institutions\_2018  & -1.10e4 & 4.85e3  & [-2.05e4, -1.51e3]       & *   \\
 & Avg\_Institutions\_2017  & -1.13e4 & 5.60e3  & [-2.23e4, -315]          & *   \\
 & Total\_Refs\_2016        &   0.265 & 0.134   & [0.00, 0.530]            & *   \\
\midrule
\multirow{3}{*}{Quartile}
 & Publication\_Delay\       & 2.10e-3 & 8.00e-4 & [5.00e-4, 3.70e-3]       & ** \\
 & AI\_Perc\_By\_LLM\_2018  & 0.728   & 0.345   & [0.052, 1.40]            & *  \\
 & AI\_Perc\_By\_LLM\_2017  & -0.929  & 0.461   & [-1.83, -0.025]          & *  \\
\bottomrule
\end{tabular}%
 }
\label{tab:reg_j1}
\vspace{-0.6cm}
\end{table}
\FloatBarrier

\begin{table*}[!htbp]
\caption{Linear mixed–effects summary for \textbf{\dataset~2023}. Significant predictors only; robust Std. Errors in parentheses.}
\centering
\setlength{\tabcolsep}{6pt}
\resizebox{\columnwidth}{!}{%
\begin{tabular}{l l r r c c}
\toprule
\textbf{Target} & \textbf{Variable} & \textbf{Coef.} & \textbf{Std. Err.} & \textbf{95\% CI (L--H)} & \textbf{Signif.} \\
\midrule
\multirow{2}{*}{Impact Factor}
 & Avg\_Authors\_2021 & 0.452 & 0.124 & [0.209, 0.696] & *** \\
 & Avg\_Authors\_2022 & 0.202 & 0.102 & [1.00e-3, 0.402] & * \\
\midrule
\multirow{5}{*}{Total Cites}
 & Std\_Authors\_2020       & 4.59e3 & 802  & [3.02e3, 6.16e3] & *** \\
 & publication\_count\_2022 & 114    & 22.6 & [70.0, 158.4]    & *** \\
 & Std\_Authors\_2022       & 3.10e3 & 867  & [1.40e3, 4.80e3] & *** \\
 & Author\_Retains          & -4.50e4 & 1.26e4 & [-6.97e4, -2.02e4] & *** \\
 & Avg\_Authors\_2021       & 1.03e4 & 4.55e3 & [1.38e3, 1.92e4] & *   \\
\midrule
\multirow{1}{*}{Quartile}
 & AI\_Perc\_By\_LLM\_2020  & 0.240  & 0.093 & [0.058, 0.421]   & ** \\
\bottomrule
\end{tabular}%
}
\label{tab:reg_j2}
\vspace{-0.5cm}
\end{table*}

We estimate all models using Restricted Maximum Likelihood (REML), which yields approximately unbiased variance–component estimates in mixed models~\cite{patterson1971reml}, and optimize the likelihood via L-BFGS, a limited-memory quasi-Newton method well suited to high-dimensional fixed effects~\cite{byrd1995limited}. The reported fixed-effect coefficients \((\boldsymbol{\beta})\) represent conditional associations within subject categories, controlling for all other covariates and random effects. We report 95\% confidence intervals and display in Tables~\ref{tab:reg_j1}--\ref{tab:reg_j2} only covariates that are statistically significant at the \(p<0.05\) level. We use asterisks to indicate significance levels: \(*\,p<0.05\), \(**\,p<0.01\), and \(***\,p<0.001\).

Across both periods, the fixed effects reveal stable yet evolving relationships between collaboration intensity, referencing practices, and AI engagement on journal-level outcomes.

For \emph{Impact Factor}, collaboration indicators consistently exhibit strong positive associations. In \dataset~2019, the average number of authors per paper (\texttt{Avg\_Authors\_2016}) had a sizable and statistically significant effect (\(\beta=0.81,\,p<0.001\)), implying that journals fostering larger research teams tend to achieve higher citation-based impact. This pattern persisted in \dataset~2023, though attenuated in magnitude, suggesting a saturation effect as multi-author collaboration became standard across fields. The standard deviation of institutional counts (\texttt{Std\_Institutions\_2018}) was negatively associated with impact, indicating that excessive institutional heterogeneity may dilute coordination efficiency or research coherence. Notably, while collaboration metrics significantly influence the \emph{Impact Factor}, they exhibit no significant relationship with \emph{Quartile} outcomes (see also Fig.~\ref{fig:quartile-collaboration-grid}, panel~d).

For \emph{Total Cites}, temporal and cross-variable effects show a complex structure. 
Earlier publication volumes (\texttt{publication\_count\_2016}) are negatively associated 
with subsequent citation totals, while more recent volumes in 2018 show strong positive effects.  Reference-related variables (\texttt{Total\_Refs}) also show alternating signs across years, 
implying that citation density does not uniformly translate into higher total citation counts 
once other factors are controlled. 
Indicators of collaboration diversity, particularly variation in author and institutional 
participation, are positively associated with total citation counts, suggesting that greater 
heterogeneity in research teams corresponds to wider citation visibility.

For the \emph{Quartile} outcome, AI-related indicators are statistically significant in both estimation periods. In \textbf{\dataset~2019}, the AI engagement rate in 2017 and 2018 display coefficients of opposite sign, indicating that early fluctuations in the share of AI-related content were not yet systematically linked to journal ranking. By \textbf{2023}, the coefficient for \texttt{AI\_Perc\_By\_LLM\_2020} is positive and statistically significant (\(\beta = 0.24,\, p < 0.01\)), showing that journals with higher proportions of AI-focused publications are more likely to occupy higher quartile positions. This shift underscores the increasing integration of AI methodologies into the core of biomedical research and their growing association with higher journal prestige.

\section{Evaluation on LLM-based Feature Extraction}

\subsection{Quality Evaluation on Select Journal Categories}

\begin{figure}[htbp!]
  \centering
  \vspace{-0.5cm} \includegraphics[width=0.31\textwidth]{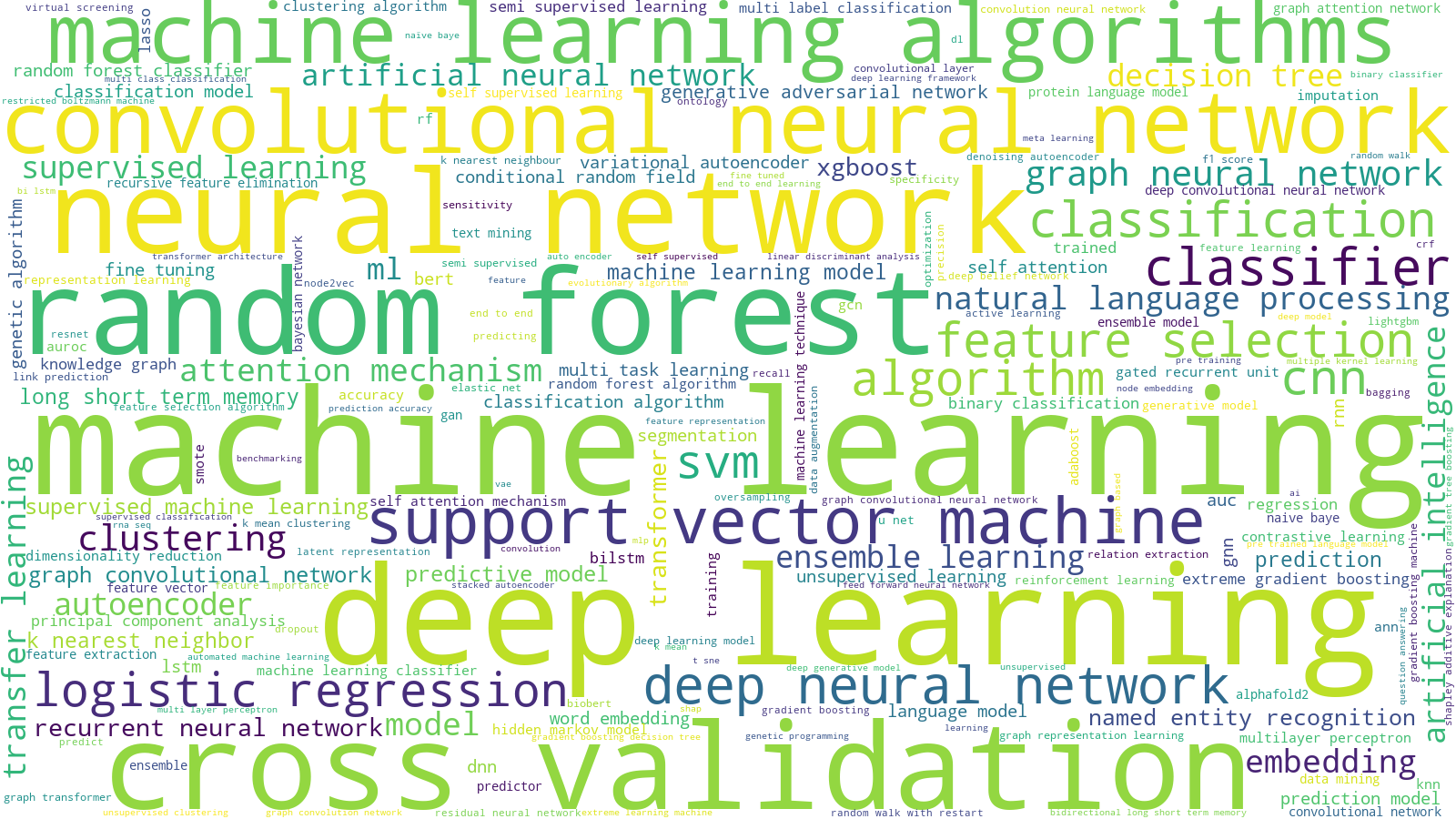}
  \includegraphics[width=0.31\textwidth]{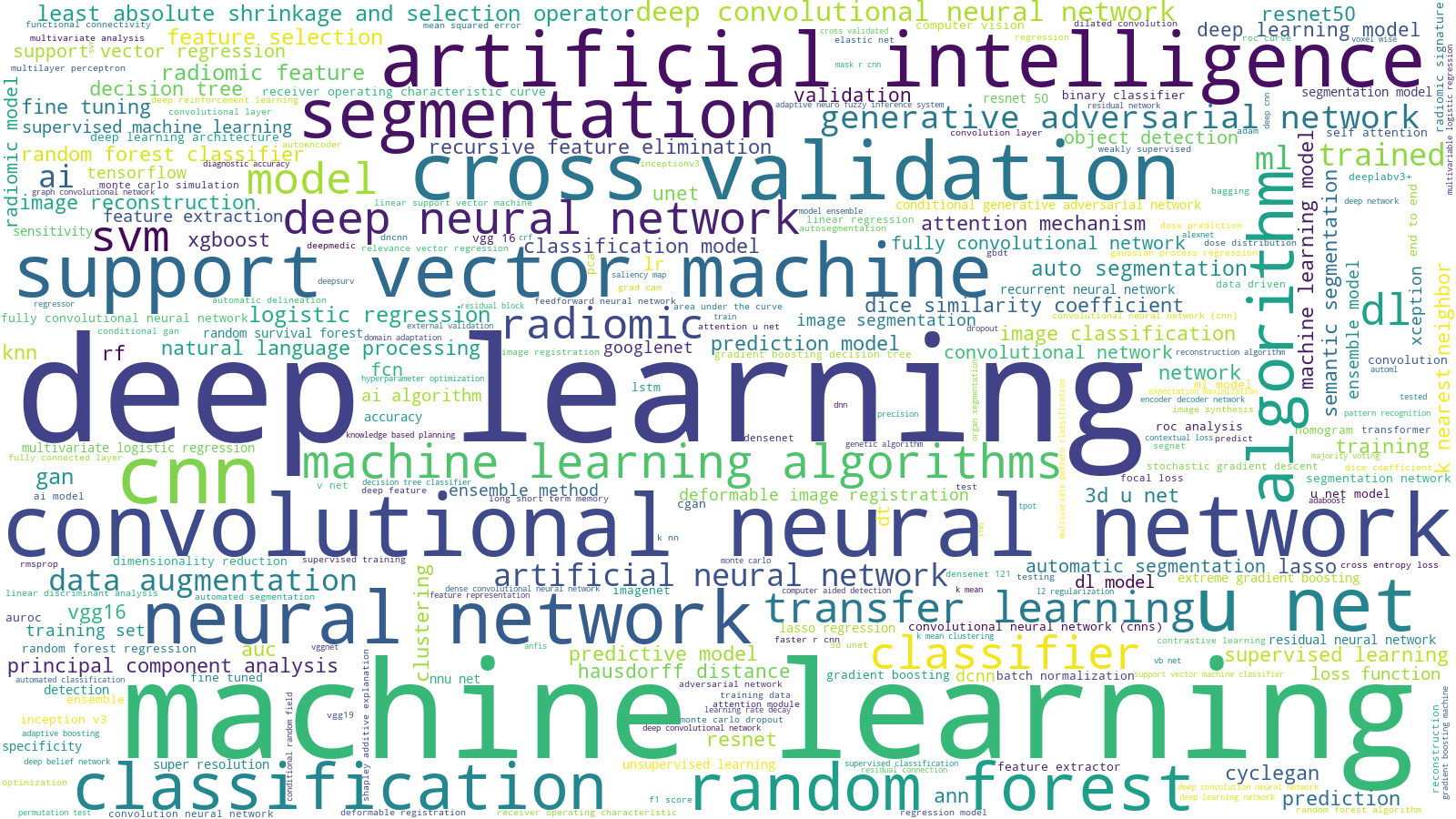}
  \includegraphics[width=0.31\textwidth]{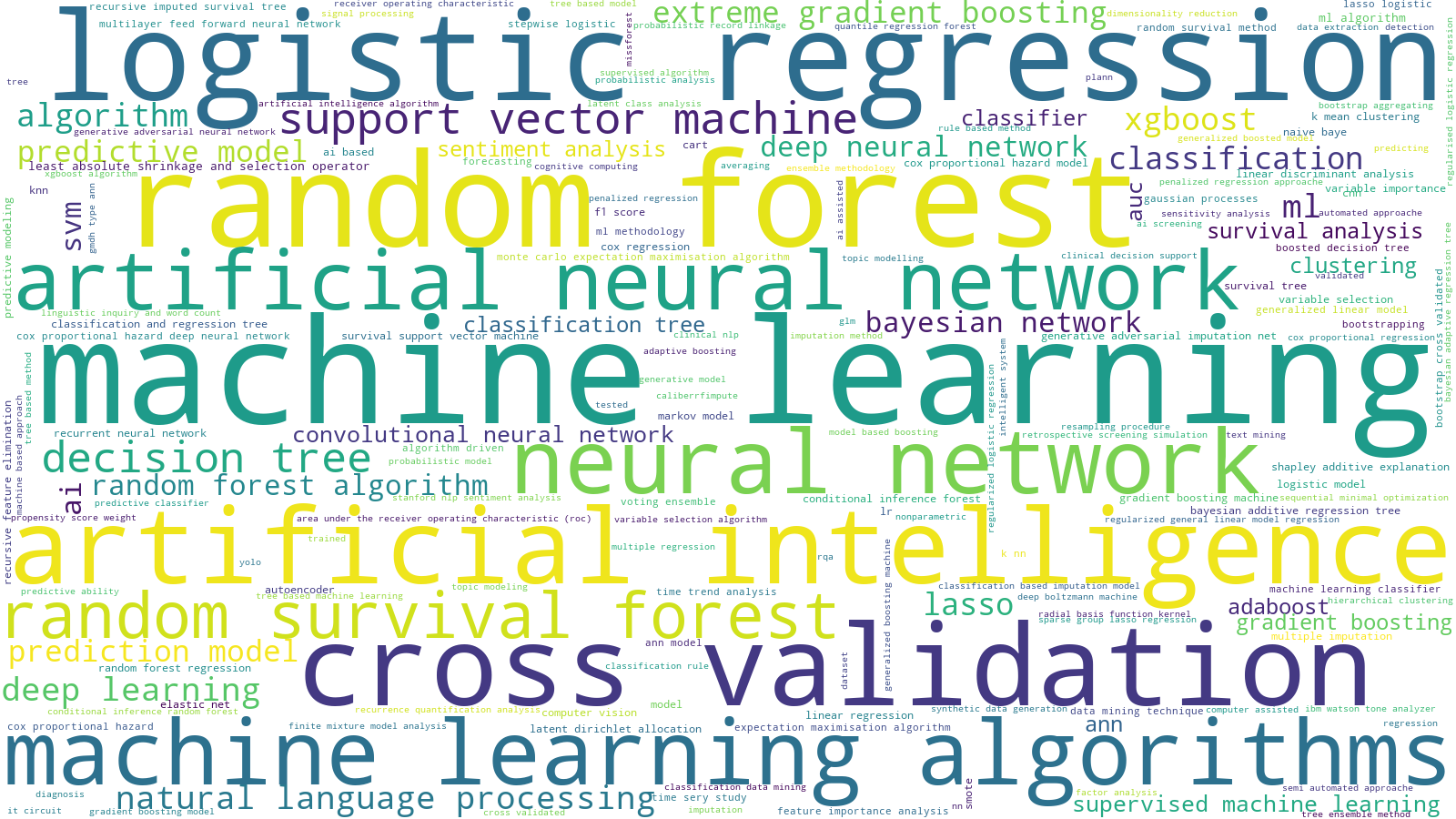}
  \caption{Subject category-specific word clouds of validated AI subfield keywords. 
  From left to right: (a) Math and Computational Biology, (b) Radiology and Imaging, and (c) Healthcare Science and Services. 
  Word size reflects frequency of extracted AI concepts within each journal subset; color and position are aesthetic only.}
\label{fig:subfieldwordclouds}
\vspace{-0.5cm} 
\end{figure}

To visualize the semantic landscape detected by the model and highlights the most prevalent AI concepts characterizing each disciplinary field, we generate \emph{word clouds} using AI keywords from Step~2 (Keyword Extraction and Subfield Mapping), validated by Step~3 (Validation Gate) of the LLM pipeline (Figure~\ref{fig:subfieldwordclouds}). Specifically, for each \emph{subject category} (e.g., Math and Computational Biology, Radiology and Imaging), we aggregate all validated AI-related keywords extracted from journals belonging to that subject category, and the word cloud is then generated by computing the normalized frequency of each keyword within the subject category:
\begin{equation}
f(w) = \frac{n(w)}{\sum_{w'} n(w')},
\end{equation}
where \(n(w)\) is the count of keyword \(w\) across all AI-relevant articles in that discipline. Word size in the cloud reflects the relative frequency of the keyword, while color and layout are aesthetic only. 

From the word clouds, Math and Computational Biology journals prominently feature a blend of classical machine learning and deep learning approaches. Frequently occurring terms such as machine learning, deep learning, neural network, random forest, and cross-validation suggest a strong focus on general-purpose predictive modeling and model evaluation. Deep learning architectures like convolutional neural networks also appear, alongside references to graph neural networks, reflecting applications to structured biological data such as molecular graphs and protein interaction networks. Radiology and Imaging journals are strongly dominated by deep learning and image-based architectures. Terms such as convolutional neural network, U-Net, classification, segmentation, and support vector machine appear frequently, indicating the prevalence of supervised computer-vision  tasks. Compared to Math and Computational Biology, tree-based models like random forest and linear models such as logistic regression are less emphasized, consistent with the field’s emphasis on imaging rather than tabular data.
In contrast, Healthcare Science and Services place greater emphasis on interpretable and clinically aligned models. Prominent terms include logistic regression, random forest, machine learning, and cross-validation, suggesting a methodological focus aligned with electronic health records, claims data, and decision-support settings where transparency, robustness, and reproducibility are prioritized over complex image-based architectures.
Across all three domains, machine learning and deep learning act as shared methodological foundations, but their use varies by data and application: imaging studies center on convolutional neural networks and segmentation tasks, biological sciences combine statistical learning with representation learning, and healthcare services emphasize interpretable models for decision-making. These trends should be viewed cautiously. We classify AI subfields using the ACM CCS system, which, though widely used, is not tailored to biomedical research and can group diverse work under broad categories like machine learning algorithms. While we partially address this by extracting AI-related keywords from titles and abstracts, our focus was on accurate category assignment rather than exhaustive keyword validation. As such, some specialized tasks or emerging methods may be underrepresented in the word clouds.

\subsection{Human Evaluation of LLM-Based Annotations}
\begin{figure*}[htbp!]
\vspace{-0.5cm} 
\centering
\begin{tabular}{cc}
    \includegraphics[width=0.49\textwidth]{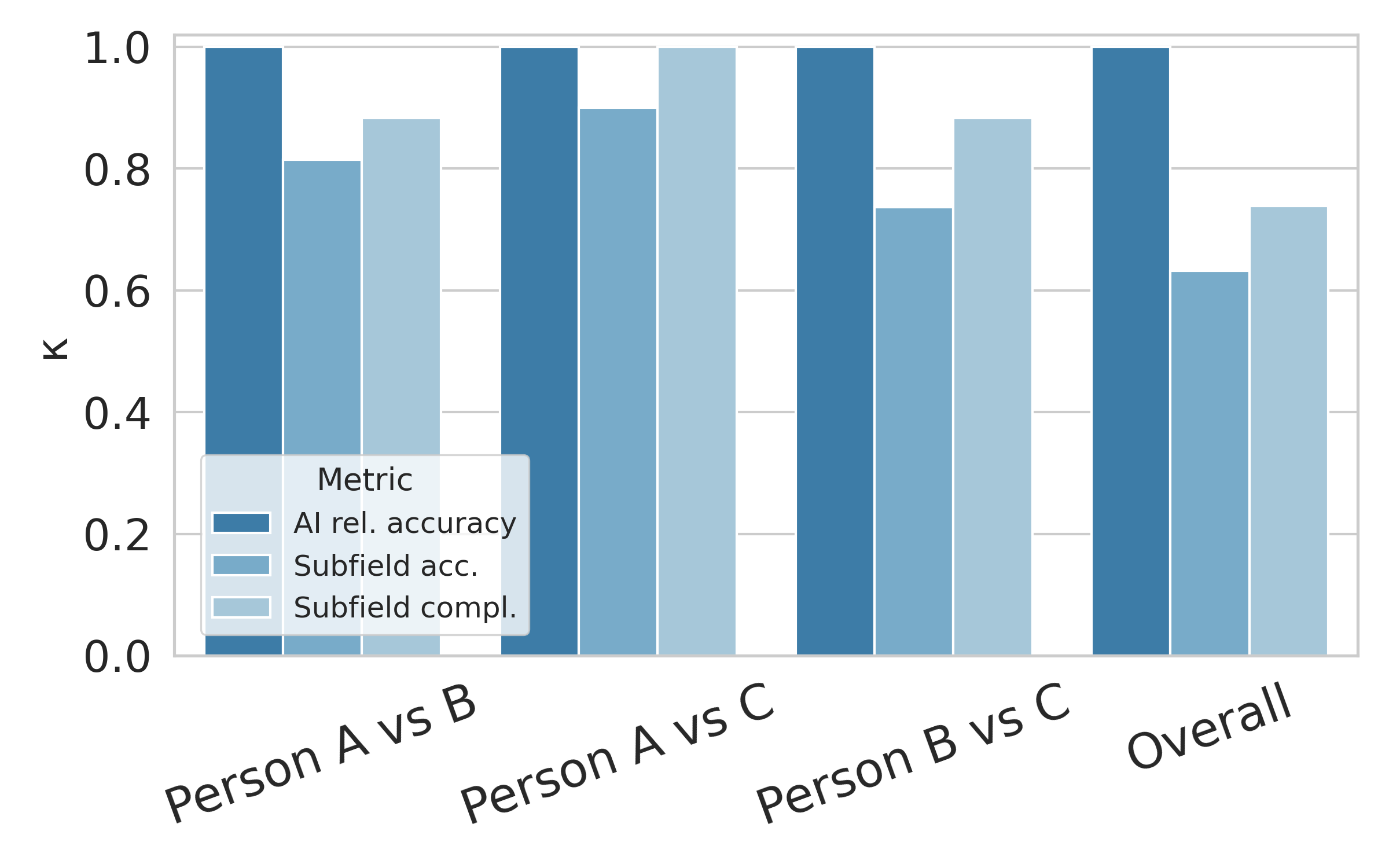} &
    \includegraphics[width=0.47\textwidth]{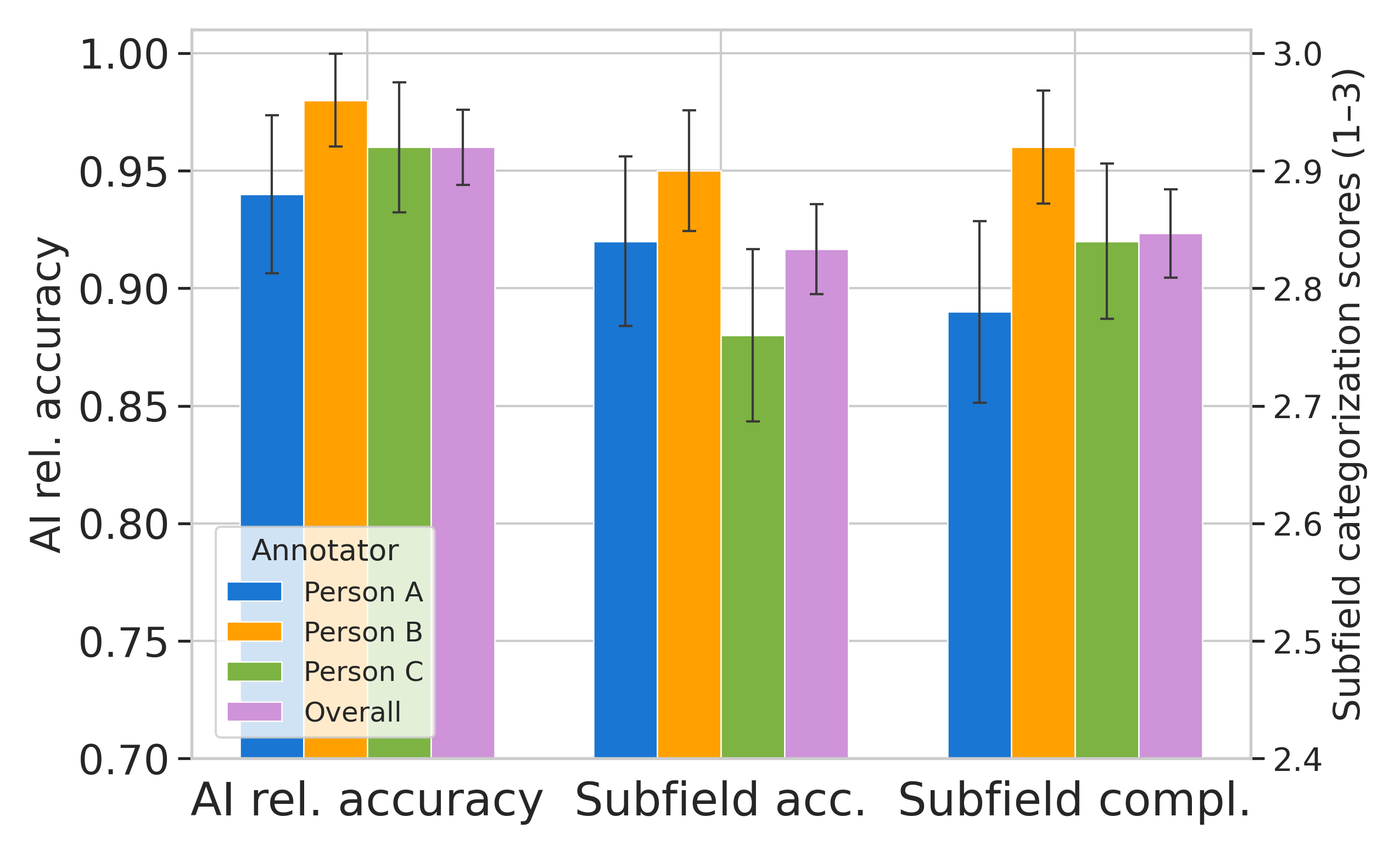} \\[-2pt]
    (a) Agreement ($\kappa$) & (b) Mean scores with Std error
\end{tabular}
\caption{Human evaluation results. (a) Pairwise Cohen's $\kappa$ for each annotator pair and metric, with an ``Overall'' bar showing three-rater Fleiss' $\kappa$.
(b) Per-annotator and overall mean scores with standard errors: AI relevance accuracy (left axis, 0–1) and subfield accuracy/completeness (right axis, 1-3).}
\label{fig:annotation-grid}
\vspace{-0.5cm}
\end{figure*}

To assess the reliability of the LLM-based annotation pipeline, we conduct a controlled human evaluation on 100 biomedical articles from PMC. Three annotators (denoted A, B, and C) participate in the study. The articles are divided into four sets of 25: one set is annotated by all three annotators, and each of the remaining three sets is assigned exclusively to one annotator, resulting in 50 annotations per annotator and 25 shared articles for every annotator pair. Each article is evaluated along three dimensions:

\begin{itemize}[leftmargin=10pt]
    \item \textbf{AI relevance accuracy}: whether the abstract explicitly involves artificial intelligence or machine learning.
    \item \textbf{Subfield accuracy (1-3)}: how accurately LLM-identified AI subfields or keywords reflect the actual content based on ACM CCS system(1 = incorrect, 2 = partially correct, 3 = completely correct).
    \item \textbf{Subfield completeness (1-3)}: whether the extracted AI subfields cover the key AI-related technical aspects in the abstract based on ACM CCS system (1 = insufficient, 2 = partial, 3 = fully complete).
\end{itemize}

We evaluate the reliability of LLM-generated annotations using $\kappa$ statistics, which account for agreement beyond chance. Pairwise agreement between annotators is measured using Cohen’s $\kappa$, while Fleiss’ $\kappa$ is reported for the subset of 25 abstracts annotated by all three annotators. Since downstream subfield judgments are only meaningful when AI content is correctly detected, if the LLM misclassifies AI relevance, the corresponding subfield accuracy and completeness scores are set to 1 by design.

Figure~\ref{fig:annotation-grid}(a) shows that AI relevance annotation reaches substantial agreement across annotators ($\kappa > 0.8$), demonstrating that the LLM’s AI detection is highly reproducible. Subfield accuracy and completeness show moderate to substantial agreement ($\kappa > 0.6$), indicating higher subjectivity in assessing technical precision and coverage, yet still reflecting consistent human interpretation. Figure~\ref{fig:annotation-grid}(b) reports the individual and overall scores for each annotation dimension. AI relevance accuracy remains consistently high across annotators, while slight variations in subfield completeness indicate that some annotators apply stricter criteria for assessing coverage. Overall, the LLM-based pipeline demonstrates strong performance across all evaluation dimensions, with strong ability in AI relevence detection especially. These results validate the LLM pipeline’s design as a reliable foundation for \dataset, demonstrating its ability to consistently extract meaningful AI-related information from biomedical abstracts and to enable scalable, interpretable, large-scale content-level analysis of scientific impact.
\section{Conclusion}
% Now your logic of the conclusion:
% you build a dataset -> you use a LLM pipeline -> Our dataset is great -> we do AI engagement analysis and human evaluation -> we see two trends -> Overall, we find LLm is a scalable tool for semantic analysis
% Do you find the problem of logic? It is messy. 
% A better logic: you build a dataset -> this dataset is really huge with biomedical journals -> this dataset also analyze the AI engagement and we find trends -> besides, we find LLM pipeline is useful -> Finally, our dataset have huge application values that can boost ...
% You can define your clear logic and put it in the comment area before you construct any paragraphs in academic writing, this will make you and your co-authors track your logic easier

We introduced \textbf{BioMedJImpact}, a large-scale, biomedical-oriented dataset that advances the study of journal-level scientific impact and AI engagement. Built from over 1.7 million PMC articles across 2,700 journals, \dataset\ integrates bibliometric indicators, collaboration indicators, and LLM-derived AI-related semantic content indicators, providing a unified resource for understanding how biomedical publishing evolves in the AI era. Using this dataset, we examine how collaboration and AI engagement jointly shape scientific impact. Two consistent trends emerge: journals with greater collaboration intensity achieve higher citation impact, and AI engagement has become an increasingly strong correlate of journal prestige, particularly in quartile rankings. The underlying LLM-based annotation pipeline is further validated through human evaluation, confirming substantial agreement in AI relevance detection and consistent subfield classification. Together, these results demonstrate that \textbf{BioMedJImpact} offers both a comprehensive dataset capturing the intersection of biomedicine and AI and a reliable, scalable methodology for content-aware scientometric analysis—supporting future research on scientific impact, innovation, and the evolving role of AI in scholarly publishing.

% \begin{credits}
% \subsubsection{\ackname} A bold run-in heading in small font size at the end of the paper is
% used for general acknowledgments, for example: This study was funded

% \subsubsection{\discintname}
% It is now necessary to declare any competing interests or to specifically
% state that the authors have no competing interests. Please place the
% statement with a bold run-in heading in small font size beneath the
% (optional) acknowledgments\footnote{If EquinOCS, our proceedings submission
% system, is used, then the disclaimer can be provided directly in the system.},
% for example: The authors have no competing interests to declare that are
% relevant to the content of this article. Or: Author A has received research
% grants from Company W. Author B has received a speaker honorarium from
% Company X and owns stock in Company Y. Author C is a member of committee Z.
% \end{credits}
\bibliographystyle{splncs04}
% Point to your .bib file (without extension). You can have multiple, comma-separated.
\bibliography{references}

\end{document}